\pdfoutput=1
\documentclass[sigconf]{acmart}

\definecolor{gray}{RGB}{192,192,192}

\usepackage{smartdiagram}
\usepackage{tikz}
\usepackage{pgfplots}
\pgfplotsset{compat=1.17}
\usetikzlibrary{
    patterns,
}

\usepackage{booktabs}
\usepackage{bm}
\usepackage{caption}
\usepackage{subfigure}
\usepackage{color}
\usepackage{multirow}

\definecolor{MyPink}{RGB}{255,178,178}
\definecolor{MyBlue}{RGB}{178,178,255}

\usetikzlibrary{er,positioning}

\copyrightyear{2021}
\acmYear{2021} 
\setcopyright{iw3c2w3}
\acmConference[WWW '21]{Proceedings of the Web Conference 2021}{April 19--23, 2021}{Ljubljana, Slovenia} 
\acmBooktitle{Proceedings of the Web Conference 2021 (WWW '21), April 19--23, 2021, Ljubljana, Slovenia}
\acmPrice{}
\acmDOI{10.1145/3442381.3449859}
\acmISBN{978-1-4503-8312-7/21/04}

\begin{document}

\title{Efficient Non-Sampling Knowledge Graph Embedding}

\settopmatter{authorsperrow=4}

\author{Zelong Li$^\ast$}
\thanks{$^\ast$The first two authors contributed equally to the work.}
\affiliation{
  \institution{Rutgers University}
  \city{New Brunswick, NJ, US}
}
\email{zelong.li@rutgers.edu}

\author{Jianchao Ji$^\ast$}
\affiliation{
  \institution{Rutgers University}
  \city{New Brunswick, NJ, US}
}
\email{jianchao.ji@rutgers.edu}

\author{Zuohui Fu}
\affiliation{
  \institution{Rutgers University}
  \city{New Brunswick, NJ, US}
}
\email{zuohui.fu@rutgers.edu}

\author{Yingqiang Ge}
\affiliation{
  \institution{Rutgers University}
  \city{New Brunswick, NJ, US}
}
\email{yingqiang.ge@rutgers.edu}

\author{Shuyuan Xu}
\affiliation{
  \institution{Rutgers University}
  \city{New Brunswick, NJ, US}
}
\email{shuyuan.xu@rutgers.edu}

\author{Chong Chen}
\affiliation{
  \institution{Tsinghua University}
  \city{Beijing, China}
}
\email{cc17@mails.tsinghua.edu.cn}

\author{Yongfeng Zhang}
\affiliation{
  \institution{Rutgers University}
  \city{New Brunswick, NJ, US}
}
\email{yongfeng.zhang@rutgers.edu}

\def\authors{Zelong Li, Jianchao Ji, Zuohui Fu, Yingqiang Ge, Shuyuan Xu, Chong Chen, Yongfeng Zhang}

\begin{abstract}
Knowledge Graph (KG) is a flexible structure that is able to describe the complex relationship between data entities. Currently, most KG embedding models are trained based on negative sampling, i.e., the model aims to maximize some similarity of the connected entities in the KG, while minimizing the similarity of the sampled disconnected entities. Negative sampling helps to reduce the time complexity of model learning by only considering a subset of negative instances, which may fail to deliver stable model performance due to the uncertainty in the sampling procedure. To avoid such deficiency, we propose a new framework for KG embedding---Efficient Non-Sampling Knowledge Graph Embedding (NS-KGE). The basic idea is to consider all of the negative instances in the KG for model learning, and thus to avoid negative sampling. The framework can be applied to square-loss based knowledge graph embedding models or models whose loss can be converted to a square loss.
A natural side-effect of this non-sampling strategy is the increased computational complexity of model learning. To solve the problem, we leverage mathematical derivations to reduce the complexity of non-sampling loss function, which eventually provides us both better efficiency and better accuracy in KG embedding compared with existing models. Experiments on benchmark datasets show that our NS-KGE framework can achieve a better performance on efficiency and accuracy over traditional negative sampling based models, and that the framework is applicable to a large class of knowledge graph embedding models.

\end{abstract}
\balance
%
%



\maketitle

\section{Introduction}
\label{sec:introduction}
Nowadays, Knowledge Graph (KG) is an important structure to store and process structured information, and has been widely used as an advanced knowledge interlinking solution in many applications. Concretely, KG is a collection of interlinked descriptions of entities. For example, Freebase, which is regarded as a practical and massive tuple database used to structure general human knowledge, is powered by KG algorithms \cite{bollacker2008freebase}. Yet another Great Ontology (YAGO) also benefits from KG structures when building its light-weight and extensible ontology with high coverage and quality \cite{suchanek2007yago}. Moreover, DBpedia builds a large-scale, multilingual knowledge base by extracting structured data from Wikipedia editions in 111 languages \cite{lehmann2015dbpedia}. These knowledge bases power a wide range of intelligent systems in practice.

Although KG has been proved as an effective method to represent large-scale heterogeneous data \cite{2020survey}, it suffers from high computation and space cost when searching and matching the entities in a discrete symbolic space. In order to leverage the power of KG more efficiently, researchers have proposed Knowledge Graph Embedding (KGE), which represents and manipulates KG entities in a latent space \cite{bordes2013translating}. 
In particular, KGE techniques embed the components of KG, including entities and relations, into a continuous vector space, so as to simplify the manipulation while preserving the inherent structure of the KG \cite{embedding}. With the help of KGE, implementing knowledge graph operations to a large scale becomes practical.

Over the past few years, a lot of efforts have been put into developing embedding algorithms, and a growing number of embedding models are proven effective. 
Most of the current embedding methods, however, depend on a very basic operation in model training called negative sampling \cite{rendle2012bpr,bordes2013translating}, which randomly or purposely samples some disconnected entities as negative samples (compared to the connected entities as positive samples), and the embedding model aims to distinguish positive vs. negative samples in the loss function for embedding learning. Typical examples include DistMult \cite{DistMult}, SimplE \cite{kazemi2018simple}, ComplEx \cite{trouillon2016complex}, TransE \cite{bordes2013translating}, RESCAL \cite{RESCAL}, etc. 
While negative sampling increases the training efficiency, it also brings several drawbacks in model accuracy. 
On one hand, only considering part of the negative instances weakens the prediction accuracy of the learned embeddings. 
And on the other hand, it also makes model training unstable because the sampled negative instances may vary in different runs. Previous studies have shown that negative sampling keeps bringing fluctuations since the result highly relies on the selection of negative samples \cite{wang2018modeling}, and these fluctuations cannot be removed by doing more updating steps \cite{chen2019efficient}. Some research tried to overcome this problem by using carefully designed sampling strategies instead of random sampling \cite{zhang2019nscaching,cai2018kbgan}. However, all of these models can only consider part of the information from the training dataset due to negative sampling.
 
Inspired by recent progress on non-sampling recommendation and factorization machines \cite{rendle2010factorization,rendle2012factorization,chen2020context,chen2020efficient,chen2020efficient2,chen2020jointly}, we make an attempt to apply the non-sampling approach to KGE. We propose a Non-Sampling Knowledge Graph Embedding (NS-KGE) framework.
The framework can be applied to square loss based KGE models. 
To apply this framework to models with other loss functions, we need to transform the loss function to square loss. 
 The framework allows us to skip the negative sample selection process and thus take all of the positive and negative instances into consideration when learning the knowledge graph embeddings, which helps to increase the embedding accuracy. A problem that naturally arises from this strategy is
the time and space complexity increase dramatically when considering all instances for embedding learning. To solve the problem,
we offer a mathematical derivation to re-write the non-sampling loss function by dis-entangling the interaction between entities, which gives us better time and space complexity without sacrificing of mathematical accuracy.
Eventually, non-sampling based KGE achieves better prediction accuracy with similar space and much shorter running time than existing negative sampling based KGE models.
To evaluate the performance of our NS-KGE framework, we apply the framework on four KGE models, including DistMult \cite{DistMult}, SimplE \cite{kazemi2018simple}, ComplEx \cite{trouillon2016complex}, and TransE \cite{bordes2013translating}. Experimental results show that the NS-KGE framework outperforms most of the models in terms of both prediction accuracy and learning efficiency. 

This paper makes the following key contributions:
\begin{itemize}
    \item We propose a Non-Sampling Knowledge Graph Embedding (NS-KGE) framework for learning effective knowledge graph embeddings.
    \item We derive an efficient method to mitigate the time and space bottlenecks caused by the non-sampling strategy.
    \item We demonstrate how the framework can be mathematically applied to existing KGE models by using DistMult \cite{DistMult}, SimplE \cite{kazemi2018simple}, ComplEx \cite{trouillon2016complex}, and TransE \cite{bordes2013translating} as examples.
    \item We conduct comprehensive experiments---including both quantitative and qualitative analyses---to show that the framework increases both accuracy and efficiency for knowledge graph embedding.
\end{itemize}

In the following part of this paper, we first introduce the related work in Section \ref{sec:related_work}. 
In Section \ref{sec:framework}, we introduce our NS-KGE framework in detail, and in Section \ref{sec:application} we show how the framework can be applied to different KGE models. 
We provide and analyze the experimental results in Section \ref{sec:experiment}, 
and conclude the work together with future directions in Section \ref{sec:conclusions}.

\section{Related work}
\label{sec:related_work}
During recent years, Knowledge Graph Embedding (KGE) has prevailed in the field of huge structured knowledge interlinking \cite{survey}, and its effectiveness has been shown in many different scenarios such as search engine \cite{xiong2017explicit,dietz2018utilizing,ai2019explainable}, recommendation system \cite{xian2019reinforcement,ma2019jointly,xian2020cafe,fu2020fairness,wang2019unified,ai2018learning,zhang2016collaborative}, question answering \cite{baidu,lin2018multi,xiong2017deeppath}, video understanding \cite{geng2021dynamic}, conversational AI \cite{fu2020cookie,zhou2020improving} and explainable AI \cite{ai2018learning,ai2019explainable,zhang2020explainable}.

Since knowledge graph embedding has extraordinary advantages in practical applications, many KGE models have been proposed. For example, Translational Embedding (TransE) model \cite{bordes2013translating} takes vector translation on spheres to model entity relationships, Translation on Hyperplane (TransH) model \cite{wang2014knowledge} enables vector translation on hyperplanes for embedding, while Translation in Relation spaces (TransR) model \cite{lin2015learning} conducts vector translation in relation-specific entity spaces for embedding. Later, 
DistMult \cite{DistMult} uses a diagonal matrix to represent the relation between head and tail entities, and the composition of relations is characterized by matrix multiplication, while ComplEx \cite{trouillon2016complex} puts DistMult into the complex domain and uses complex numbers to represent the head-relation-tail triples in the knowledge graph.
More recently, SimplE \cite{kazemi2018simple} provides a simple  enhancement of the Canonical Polyadic (CP) \cite{hitchcock1927expression} tensor factorization for interpretable knowledge graph embedding. More comprehensive review of knowledge graph embedding techniques can be seen in \cite{survey,hogan2020knowledge}.

Most of the existing KGE models rely on negative sampling for model learning, which randomly sample some disconnected entities to distinguish with connected entities, and meanwhile reduce the training time compared with using all negative samples.
However, due to the uncertainty of sampling negative instances, the results of embedding learning may fluctuate greatly in different runs. Besides, some models only produce satisfying embedding results when the number of negative samples is large enough \cite{DistMult,trouillon2016complex}, which increases the time needed for model training.

In previous works, some methods \cite{zhang2013dynamic, wang2018incorporating} have been developed to mitigate the above problems, mostly by sampling the negative instances purposely rather than randomly. For example, 
dynamic negative sampling \cite{zhang2013dynamic} chooses negative training instances from the ranking list produced by the current prediction model, so that the model can continuously work on the most difficult negative instances. Generative Adversarial Networks (GAN) are also used to generate and discriminate high-quality negative samples \cite{wang2018incorporating}, which take advantage of a generator to obtain high-quality negative samples, and meanwhile the discriminator in GAN learns the embeddings of the entities and relations in knowledge graph so as to incorporate the GAN-based framework into various knowledge grahp embedding models for better ability of knowledge representation learning. Ultimately, these models still rely on the sampled negative instances instead of all instances for model training and the model accuracy still has room for improvement.

Recently, researchers have explored whole-data based approaches to train recommendation models \cite{chen2020efficient, chen2020efficient2, chen2020context},
which improve the recommendation accuracy without negative sampling. By separating the users and items in optimization, the computational bottlenecks has been resolved in a large extent in training the recommendation models. However, these methods can only be applied to recommendation models, while we would like to build a general framework that can be applied to square-loss-based knowledge graph embedding models. 
Besides, although these models achieve better recommendation performance and efficiency, they do not consider improving the space complexity but only focus on the time complexity, and thus they still need to use batch learning during the training process. On the contrary, we aim to improve both the space and time complexity in this work. Based on this, we can achieve three benefits: better entity ranking accuracy, better computational efficiency, and better space efficiency.

\section{Non-Sampling KGE Framework}
\label{sec:framework}
In this section, we will first introduce the notations that will be used in this paper. Then, we will introduce the Non-Sampling Knowledge Graph Embedding (NS-KGE) framework step by step. In particular, we will first provide a general formulation of the framework, and then devote two subsections to show how to improve the time and space efficiency in the framework. We will show how the framework can be applied to different specific knowledge graph embedding models in the next section.

\subsection{Problem Formalization and Notations}
In this section, we provide a square-loss based formalization of the Knowledge Graph Embedding (KGE) problem, which will be used in the following parts of the paper. However, we acknowledge that not all of the existing KGE methods can be represented by this square loss formalization. In this paper, we consider those KGE methods whose loss function is a square loss or can be converted into a square loss format for non-sampling KGE learning.
Table \ref{Table:notation} introduces the basic notations that will be used in this paper. 
We first provide a general formulation for the knowledge graph embedding problem. Given a knowledge graph $\bm{G}$, our goal is to train a scoring function $\hat{f}_r(h,t)$, which is able to distinguish whether the head entity $h$ and tail entity $t$ should be connected by relation $r$ in the knowledge graph. 
Suppose $f_r(h,t)$ is the ground-truth value of the triplet $(h,r,t)$, generated from training sets, and $\hat{f}_{r}(h,t)$ is the predicted value by the knowledge graph embedding model,
where $f_r(h,t)=1$ represents the connected entities, and $f_r(h,t)=0$ denotes dis-connected entities.
Based on these definition, a general KG embedding model aims to minimize the difference between the ground-truth and the predicted values based on a loss function ${L}$. For example, we can use square loss to train the model: 
\begin{equation}
\begin{split}
    L & = \sum_{r \in \bm R} \sum_{h \in \bm E} \sum_{t \in \bm E} c_{hrt} \left (f_{r}(h,t) -  {\hat{f}}_{r}(h,t)\right)^2
\end{split}
\label{equation:framework}  
\end{equation}
where the three summations enumerate all of the possible $(h,r,t)$ triplet combinations in the knowledge graph, and $c_{hrt}$ represents the importance (i.e., weight score) of the corresponding triplet. In traditional negative sampling-based KGE models such as TransE, $c_{hrt}=1$ is set as the positive instances and the sampled negative instances, while for all other negative instances, $c_{hrt}=0$. In our non-sampling KGE framework, however, all $c_{hrt}$ values are non-zero. In the simplest case, $c_{hrt}=1$ for all instances, regardless of positive or negative.

Many knowledge graph embedding models can be regarded as a special case of this formulation. For example, TransE uses $\hat{f}_{r}(h,t)=\|\bm e_h+\bm r-\bm e_t\|$ as the scoring function. For connected entities, the ground-truth value $f_{r}(h,t)$ would be 0, and for dis-connected entities, the ground-truth value would be a constant value greater than 0 (e.g., it would be 3 when $\bm e_h, \bm r$ and $\bm e_t$ are regularized as unit vectors).
Similar to TransE, many other KG embedding models can be represented by Eq.\eqref{equation:framework} with no or only a little trivial transformation, as we will show later in Section \ref{sec:application}.

\subsection{Non-sampling KG Embedding}
The adoption of square loss in Eq.\eqref{equation:framework} 
makes it possible to simplify the model learning and increase the time and space efficiency based on mathematical re-organization of the loss function.
In the first step, we can re-write the loss function as following:
\begin{equation}
\small
\begin{split}
    L & = \sum_{r \in \bm R} \sum_{h \in \bm E} \sum_{t \in \bm E} c_{hrt}\left (f_{r}(h,t) -  {\hat{f}}_{r}(h,t)\right )^2
    \\& = \sum_{r \in \bm R} \sum_{h \in \bm E} \sum_{t \in \bm E} c_{hrt}\left (f_{r}(h,t)^2 + \hat{f}_{r}(h,t)^2 - 2f_{r}(h,t)\hat{f}_{r}(h,t)\right )
\end{split}
    \label{equation:1}  
\end{equation}

From Eq.\eqref{equation:1}, we can see that the time complexity of calculating the loss is huge. The time complexity of calculating $\hat{f}_r(h, t)$ is $O(d)$,
where $d$ is the dimension of embedding vectors, 
and thus the time complexity of calculating the whole loss function is $O(d|\bm R||\bm E|^2)$.  If we implement this loss function on real-world knowledge graphs, we would have to conduct trillions of times of computation to calculate the loss function within one epoch. Depending on the size of the training data, this may take days, weeks or even longer to train a model. Even if using all of the (both positive and negative) samples in the dataset for model training can bring us better accuracy, such training time is not affordable. As a result, we need to mathematically derive more efficient implementations for Eq. \eqref{equation:1}.

\begin{table}[t]
    \centering
    \begin{tabular}{c|l}
    \toprule
      {\bfseries Symbol} &{\bfseries Description}\\ 
      \midrule
      $\bm G$ & A knowledge graph\\
      $\bm E$ & The set of entities in a knowledge graph\\
      $\bm R$ & The set of relations in a knowledge graph\\
      $h,t$ & A head ($h$) or a tail ($t$) entity in a knowledge graph\\
      $r$ & A relation in a knowledge graph \\
      $\bm e_h, \bm e_t$ & Embedding vector of the entity $h$ and $t$\\
      $\bm r$ & Embedding vector of the relation $r$ \\
      $e_{h,i},e_{t,i}$ & The $i$-th dimension of entity embedding $\bm e_h$ and $\bm e_t$\\
      $r_i$ & The $i$-th dimension of relation embedding $\bm r$\\
      $d$  &Dimension of the embedding vectors\\
      $c_{hrt}$ & The weight of the triplet $(h, r, t)$ \\
      $f_{r}(h,t)$ & Ground-truth value of the triple $(h, r, t)$\\
      ${\hat{f}}_{r}(h,t)$ & Predicted value of the triple $(h, r, t)$\\
      \bottomrule
    \end{tabular}
    \caption{Summary of the notations in this work.}
    \vspace{-20pt}
    \label{Table:notation}
\end{table}

\subsection{Improving Time Efficiency}
\label{section: efficiency}

To reduce the time complexity, the first thing we need to consider is to find out the most time-consuming part of the loss function. Without loss of generality and to simplify the model computation, we assume ground-truth value $f_{r}(h, t)=1$ for positive instances and $f_{r}(h, t)=0$ for negative instances. Besides, we set a uniform coefficient $c^+$ for all positive instances and $c^-$ for all negative instances. In this case the loss function can be re-organized as:
\begin{equation}
\small
\hspace{-10pt}
\begin{split}
    L & = \sum_{r \in \bm R} \sum_{h \in \bm E} \sum_{t \in \bm E} c_{hrt}\big(f_{r}(h,t) - \hat{f}_{r}(h,t)\big)^2
    \\& \overset{1}{=} \sum_{r \in \bm R} \sum_{h \in \bm E} \Big[\sum_{t \in \bm E^+_{h,r}} c^+\Big(f_{r}(h,t)^2 + \hat{f}_{r}(h,t)^2 - 2f_{r}(h,t)\hat{f}_{r}(h,t)\Big)
    \\& + \sum_{t \in \bm E^-_{h,r}}c^-\hat{f}_{r}(h,t)^2\Big]
    \\& \overset{2}{=} \sum_{r \in \bm R} \sum_{h \in \bm E} \Big[\sum_{t \in \bm E^+_{h,r}} c^+\Big(f_{r}(h,t)^2 + \hat{f}_{r}(h,t)^2 - 2f_{r}(h,t)\hat{f}_{r}(h,t)\Big) 
    \\& + \Big(\sum_{t \in \bm E}c^-\hat{f}_{r}(h,t)^2 - \sum_{t \in \bm E^+_{h,r}}c^-\hat{f}_{r}(h,t)^2\Big)\Big]
    \\& \overset{3}{=} \underbrace{\sum_{r \in \bm R} \sum_{h \in \bm E} \sum_{t \in \bm E^+_{h,r}} \Big[c^+\left(\hat{f}_{r}(h,t)^2 - 2f_{r}(h,t)\hat{f}_{r}(h,t)\right)-c^-\hat{f}_{r}(h,t)^2\Big]}_{L^P}
    \\& + \underbrace {\sum_{r \in \bm R} \sum_{h \in \bm E}\sum_{t \in \bm E}c^-\hat{f}_{r}(h,t)^2}_{L^A}+\underbrace{\sum_{r \in \bm R} \sum_{h \in \bm E}\sum_{t \in \bm E^+_{h,r}}c^+f_{r}(h,t)^2}_{\text{constant}}
\end{split} 
    \label{equation:key}
\end{equation}
where $\bm E^+_{h,r}$ represents the set of entities in the KG that are connected to head entity $h$ by relation $r$, while $\bm E^-_{h,r}$ is the set of entities that are not connected to $h$ through $r$. We have $\bm E^+_{h,r}\cup\bm E^-_{h,r}=\bm E$. In the step 1 above, we split the loss function by considering $\bm E^+_{h,r}$ and $\bm E^-_{h,r}$ separately. In step 2, we replace the $\bm E^-_{h,r}$ term by subtracting $\bm E^+_{h,r}$ from the total summation, and in step 3, we re-organize the loss function into the positive term $L^P$, all entity term $L^A$, and a constant value term. In the loss function, we separate the positive entities and the constant value, and we use all of the data to replace the negative samples. In the next step, we will introduce how to optimize the time complexity after such transformation.

As we can see, the $L^P$ term enumerates over all of the connected triplets in the KG, and its time complexity is $O(d\times$\# positive triples$)$, which is an acceptable complexity. However, since most of the KG datasets are highly sparse, $L^A$, even with pretty concise form, contributes the most significant time complexity to the loss function. Actually, the time complexity of $L^A$ is  $O(d|\bm R||\bm E|^2)$, which is very expensive. We hope the time complexity of $L^A$ can be further reduced.
As a result, we will take a closer look at the $\hat{f}_{r}(h,t)^2$ term.

Fortunately, NS-KGE can be applied to most of the factorization-based KG embedding models. For these models, we can conduct certain transformations over the scoring function $\hat{f}_{r}(h,t)$ to reduce the time complexity. 
For a factorization-based KG embedding model, the scoring function $\hat{f}_{r}(h,t)$ can be represented as the following general formulation:
\begin{equation}
\begin{split}
    {\hat f}_{r}(h,t) = \bm e_h^T (\bm r \odot \bm e_t)= \sum_i^d e_{h,i}  r_i  e_{t,i}
\end{split}
      \label{equation:fb}
\end{equation}
where $\bm e_h$, $\bm r$ and $\bm e_t$ are the head entity embedding, relation embedding, and tail entity embedding, respectively, and the symbol $\odot$ denotes element-wise product. Besides, $e_{h,i}$, $r_i$ and $e_{t,i}$ denote the $i$-th element of the corresponding embedding vector. Since our NS-KGE framework is based on square loss, we calculate the square of $\hat{f}_{r}(h,t)$. By manipulating the inner-product operation, the square of $\hat{f}_{r}(h,t)$ can be rearranged as:
\begin{equation}
\begin{split}
    {\hat f}_{r}(h,t)^2 &=  \Big(\sum_i^d e_{h,i} r_i e_{t,i} \Big) \Big(\sum_j^d e_{h,j} r_j e_{t,j}\Big)
    \\& = \Big(\sum_i^d\sum_j^d e_{h,i} e_{h,j}\Big) \Big(\sum_i^d\sum_j^d r_i r_j\Big) \Big(\sum_i^d\sum_j^d e_{t,i} e_{t,j}\Big)
\end{split}
\label{equation:5}
\end{equation}

In this case, $\bm e_h$, $\bm r$ and $\bm e_t$ are separated from each other, and thus $\sum_i^d\sum_j^d e_{h,i} e_{h,j}$, $\sum_i^d\sum_j^d r_i r_j$ and $\sum_i^d\sum_j^d e_{t,i} e_{t,j}$ are independent from each other. Therefore, we can disentangle the parameters and calculate ${\hat f}_{r}(h,t)^2$ in a more efficient way. 
We thus apply Eq.\eqref{equation:5} to the $L^A$ term of the loss function Eq.\eqref{equation:key} and we have:
\begin{equation}
\small
\begin{split}
   \bm L &= L^P + c^-\sum_{r \in \bm R} \sum_{h \in \bm E}\sum_{t \in \bm E} \Big(\sum_i^d\sum_j^d e_{h,i} e_{h,j} \Big) \Big(\sum_i^d\sum_j^d r_i r_j \Big) \Big(\sum_i^d\sum_j^d e_{t,i} e_{t,j}\Big)
   \\& = L^P + \overbrace{c^-\sum_i^d\sum_j^d \underbrace{ \Big(\sum_{h \in \bm E} e_{h,i} e_{h,j} \Big)}_{L_H} \underbrace{\Big(\sum_{r \in \bm R}  r_i r_j \Big)}_{L_R} \underbrace{\Big(\sum_{t \in \bm E} e_{t,i} e_{t,j} \Big)}_{L_T}}^{L^A}
\end{split}
      \label{equation:3.3final}
\end{equation}

In the second step of Eq.\eqref{equation:3.3final}, the reason that we can reorganize the summary of $i,j$ and the summary over $h,r,t$ is because the $L_H,L_R$ and $L_T$ terms are independent from each other. We also leave out the constant term since it does not influence the optimization result.

As noted before, $L^A$ contributes the most significant complexity to the loss function. Based on the above operation, the complexity of $L^A$ is reduced from $O(d|\bm E| |\bm R| |\bm E|)$ in Eq.\eqref{equation:key} to $O(d^2(|\bm E|+|\bm R|+|\bm E|))$ in Eq.\eqref{equation:3.3final}. In Section \ref{sec:application}, we take the Bilinear-Diagonal embedding model (DistMult), the Simple enhancement of Canonical Polyadic model (SimplE), the Complex Embedding model (ComplEx) and the Translational Embedding model (TransE) as examples to show how the NS-KGE framework can be applied to different models.

\subsection{Improving Space Efficiency}
\label{sec:space}
Apart from time complexity, we also provide a method to reduce the space complexity which is still based on the factorization-based scoring function $\hat{f}_{r}(h, t)$ in Eq.\eqref{equation:fb}. The models that will be studied in Section \ref{sec:application} satisfy this form or could be extended 
with some simple transformations.

First, we use two $|E|\times d$ matrices, $H_e$ and $T_e$, to store the embedding vectors of all head entities and tail entities, respectively. Similarly, we store the embedding vectors of all relations in the matrix $R_e$, with the size of $|R|\times d$. According to Eq.\eqref{equation:3.3final}, the calculation of
$L_H,L_R$ and $L_T$ are independent from each other with the calculation of each term only relies on the
corresponding $H_e$, $R_e$ or $T_e$ matrix. For example, given the index $i$ and $j$, the value of the $L_H$ term is equal to the inner product of the $i$-th column vector and the $j$-th column vector of matrix $H_e$. As a result, we can construct three intermediate matrices denoted as
$M_H$, $M_T$, and $M_R$ with the size $d\times d$ to record the intermediate results for each term. The details are displayed in Eq.\eqref{equation:intermediate}.
\begin{equation}
       M_H = H_{e}^TH_{e},~
       M_R = R_{e}^TR_{e},~
       M_T = T_{e}^TT_{e}
    \label{equation:intermediate}
\end{equation}

Note that $M_H[i][j]$ is equal to the inner product of the $i$-th and the $j$-th column vector of $H_e$, similar for $M_R[i][j]$ and $M_T[i][j]$. 
Based on this, the calculation of the $L^A$ term in Eq.\eqref{equation:3.3final} can be simplified as:
\begin{equation}
\label{equation:space}
    L^A = c^- sum(M_H\odot M_R\odot M_T)
\end{equation}
where $\odot$ means element-wise product of matrices, and $sum$ means adding up all elements of a matrix. In this way, we can calculate $L^A$ in the space complexity of $O \left(d \times (|R|+|E|+d) \right)$, so that we do not need to use any batch optimization for standard knowledge graph benchmarks such as the FB15K237 and WN18RR datasets (to be introduced in Section \ref{sec:experiment}), since we can directly use the whole training data to calculate the loss function within reasonable time and space complexity. 

As will be shown in the following section, for different models we may need to construct the $M_H$, $M_R$ and $M_T$ matrices in different ways, but this does not increase the space complexity. Besides, for extremely large datasets that cannot be loaded into memory as a whole, our framework with smaller space complexity can use fewer batches to train the model, which results in less training epochs.

\section{Apply NS-KGE on Different Models}
\label{sec:application}
In this section, we will show how our NS-KGE framework can be applied over different KGE models. As mentioned before, we will take the Bilinear-Diagonal embedding model (DistMult) \cite{DistMult}, the Simple enhancement of Canonical Polyadic model (SimplE) \cite{kazemi2018simple}, the Complex Embedding model (ComplEx) \cite{trouillon2016complex} and the Translational Embedding model (TransE) \cite{bordes2013translating} as examples.

\subsection{Bilinear-Diagonal Embedding (DistMult)}

DistMult is a representative factorization-based  multi-relation representation learning model \cite{DistMult}. It learns each entity as a vector embedding, and learns each relation as a diagonal matrix.
For a triplet 
$(h, r, t)$,
DistMult trains the model based on the following scoring function:
\begin{equation}
\begin{split}
    \bm{\hat f}_{r}(h,t) = \bm e^T_h \cdot diag(\bm r) \cdot \bm e_t= \sum_i^d e_{h,i} r_i e_{t,i}
\end{split}
      \label{equation:3}
\end{equation}



We can see that the scoring function of DistMult is the same as our framework (Eq.\eqref{equation:fb}). As a result, we can directly apply the loss function of the NS-KGE framework (Eq.\eqref{equation:3.3final}) for model learning, and use Eq.\eqref{equation:space} for better space complexity.
The final time complexity is $O\big(d^2(|\bm E|+|\bm R|+|\bm E|)\big)$. 
In the experiments, we will show that our Non-sampling DistMult is both more efficient and more effective that the original sampling-based DistMult model.

\subsection{Simple Enhancement of CP (SimplE)}

Canonical Polyadic (CP) decomposition is one of the earliest work on tensor factorization approaches \cite{hitchcock1927expression}. Since CP learns two independent embedding vectors for each entity, it performs poorly in KG link prediction tasks. The SimplE \cite{kazemi2018simple} embedding method provides a simple improvement of CP by learning the two embeddings of each entity dependently, which gains much better performance on link prediction.

For SimplE, its scoring function is a little bit more complex than DistMult. The scoring function is a combination of two parts:
\begin{equation}
\begin{split}
    \hat{f}_{r}(h,t) &=\frac{1}{2}\Big(\bm e_h^T(\bm r\odot \bm e_t) + \bm e_t^T(\bm r^{-1}\odot \bm e_h)\Big)
    \\ &= \frac{1}{2}\left(\sum_{i=1}^d e_{h,i}r_ie_{t,i}+\sum_{j=1}^d e_{t,j} r^{-1}_j e_{h,j}\right)
\end{split}
        \label{equation:10}
\end{equation}

To apply the NS-KGE framework on SimplE, we need to consider $\hat{f}_{r}(h,t)^2$,
which consists of three terms:
\begin{equation}
\begin{split}
    \hat{f}_{r}(h,t)^2 =\frac{1}{4}\bigg(&\sum_{i}^d\sum_{j}^d(e_{h,i} e_{h,j})(r_i r_j)(e_{t,i} e_{t,j})\\
    +&2\sum_{i}^d\sum_{j}^d(e_{h,i} e_{h,j})(r_i r_j^{-1})(e_{t,i} e_{t,j})
    \\
    +&\sum_{i}^d\sum_{j}^d(e_{h,i} e_{h,j})(r_i^{-1} r_j^{-1})(e_{t,i} e_{t,j})\bigg)
\end{split}
        \label{equation:11}
\end{equation}

We can see that for each term in Eq.\eqref{equation:11}, its structure is the same as that in Eq.\eqref{equation:5}. As a result, the rest of the work is similar to what we did in Eq.\eqref{equation:3.3final}. The only difference is that we result in three $L^A$ terms in Eq.\eqref{equation:3.3final}, but the time and space complexity remain unchanged.


\subsection{Complex Embeddings (ComplEx)}
\label{section:complex}
The ComplEx embedding model learns KG embeddings in a complex number space \cite{trouillon2016complex}. It adopts the Hermitian dot product to construct the scoring function. But we can still do similar rearrangements for non-sampling knowledge graph embedding. The scoring function of ComplEx is:
\begin{equation}
    \begin{split}
        \hat{f}_{r}(h,t) & = \bm h_{re}^T(\bm r_{re} \odot \bm t_{re}) + \bm h_{im}^T(\bm r_{re} \odot \bm t_{im}) \\
        & + \bm h_{re}^T(\bm r_{im} \odot \bm t_{im}) - \bm h_{im}^T(\bm r_{im} \odot \bm t_{re}) \\
    \end{split}
\end{equation}
where $\bm h_{re}, \bm r_{re}, \bm t_{re}$ are the real part of the head, relation, and tail embedding vectors, while $\bm h_{im}, \bm r_{im}, \bm t_{im}$ are the imaginary part of the head, relation, and tail embedding vectors. There will be six terms for $\hat{f}_{r}(h,t)^2$. In the following, we would only show the expansion result of the first term $L^A_1=\bm h_{re}^T(\bm r_{re}\odot \bm t_{re}) \cdot \bm h_{im}^T(\bm r_{re} \odot \bm t_{im})$, since the other terms look similar.
\begin{equation}
\small
    \begin{split}
    L^A_1 &= c^- \sum_{r \in \bm{R}} \sum_{h \in \bm{E}} \sum_{t \in \bm E}
    \Big(\sum_{i}^d h_{re, i} r_{re, i} t_{re, i}\Big) \Big(\sum_j^d h_{im, j} r_{re, j} t_{im, j} \Big) \\
    & = c^- \sum_i^d \sum_j^d \Big(\sum_{h \in \bm E}h_{re, i}h_{im, j}\Big) \Big(\sum_{r \in \bm R}r_{re, i}, r_{re, j}\Big) \Big(\sum_{t \in \bm E}t_{re, i}, t_{im, j}\Big) 
    \end{split}
    \label{equation:13}
\end{equation}

In this way, we can calculate the loss of non-sampling ComplEx within $O(d^2(|R|+|E|))$ time complexity and $O(d(|R|+|E|+d))$ space complexity.

\subsection{Translational Embedding (TransE)}


Unlike the previous factorization-based knowledge graph embedding models, TransE \cite{bordes2013translating} is a translation-based embedding model. As a result, the NS-KGE framework cannot be directly applied to TransE. However, we will show that by applying very minor modifications to the scoring function without changing the fundamental idea of translational embedding, the NS-KGE framework can still be applied to TransE.

The original scoring function of TransE is $\hat{f}_{r}(h, t) =\|\bm h+\bm r-\bm t\|_2^2$, which calculates the distance between $\bm h+\bm r$ and $\bm t$. For positive examples, we hope the distance would be as small as possible, which could be 0 in the most ideal case. For negative examples, we hope the distance would be as far away as possible. In many implementations of TransE, to avoid over-fitting, we usually apply unit-vector constraints to the embedding vectors, i.e., $\bm h, \bm r$ and $\bm t$ are regularized as length-one vectors. In this case, the maximum possible distance between $\bm h+\bm r$ and $\bm t$ would be 3, as a result, the optimal value for negative examples would be 3. However, the mathematical derivation of our framework in Eq.\eqref{equation:key} relies on the assumption that the ground-truth value for positive instances is 1 and for negative instances it is 0. As a result, we slightly modify the scoring function of TransE to a new function $\hat{f}^\prime_{r}(h, t) = 1 -\frac{1}{3} \hat{f}_{r}(h,t) = 1-\frac{1}{3}\|\bm h+\bm r-\bm t\|_2^2$. In this way, the ground-truth value satisfy our assumption meanwhile it does not influence the optimization of TransE. The only difference is that instead of minimizing $\hat{f}_{r}(h, t)$ for positive examples in TransE, we aim to maximize $\hat{f}^\prime_{r}(h, t)$, which is basically the same in terms of optimization.



By inserting $\hat{f}^\prime_{r}(h, t)$ into Eq.\eqref{equation:key}, we have:
\begin{equation}
\Small
    \begin{split}
        L & = L^P + \sum_{r \in \bm R} \sum_{h \in \bm E}\sum_{t \in \bm E}c^-\Big(1 -\frac{1}{3} \hat{f}_{r}(h,t)\Big)^2 \\
        & = L^P + \sum_{r \in \bm R} \sum_{h \in \bm E}\sum_{t \in \bm E} c^-\Bigg(\frac{2}{3}\Big(\bm h^T\bm t + \bm r^T(\bm t-\bm h)\Big)\Bigg)^2 \\
        & = L^P + \sum_{r \in \bm R} \sum_{h \in \bm E}\sum_{t \in \bm E} \frac{4c^-}{9}\left((\bm h^T\bm t)^2 + (\bm r^T\bm t)^2+(\bm r^T\bm h)^2-2\bm r^T\bm t\bm r^T\bm h \right) \\
        & = L^P + \frac{4c^-}{9}\Big(\sum_{i}^d \sum_{j}^d|E|\sum_{r \in \bm R} r_ir_j\sum_{t \in \bm E} t_it_j + \sum_{i}^d \sum_{j}^d|E|\sum_{h \in \bm E} h_ih_j\sum_{r \in \bm R} r_ir_j \\
        & + \sum_{i}^d \sum_{j}^d|R|\sum_{h \in \bm E} h_ih_j\sum_{t \in \bm E} t_it_j - 2\sum_{i}^d \sum_{j}^d\sum_{r \in \bm R} r_ir_j\sum_{h \in \bm E} h_i \sum_{t \in \bm E} t_j \Big) \\
        & = L^P + \frac{4c^-}{9}\sum_{i}^d \sum_{j}^d\Big(|E|\underbrace{\sum_{r \in \bm R} r_ir_j}_{L_R} \underbrace{\sum_{t \in \bm E} t_it_j}_{L_T} + |E| \underbrace{\sum_{h \in \bm E} h_ih_j}_{L_H} \underbrace{\sum_{r \in \bm R} r_ir_j}_{L_R} \\
        & + |R|\underbrace{\sum_{h \in \bm E} h_ih_j}_{L_H} \underbrace{\sum_{t \in \bm E} t_it_j}_{L_T} - 2\underbrace{\sum_{r \in \bm R} r_ir_j}_{L_R} \underbrace{\sum_{h \in \bm E} h_i}_{S_H} \underbrace{\sum_{t \in \bm E} t_j}_{S_T} \Big)
    \end{split}
    \label{equation:transe}
\end{equation}

We can see that similar to Eq.\eqref{equation:3.3final}, the final loss can also be decomposed to the $L_H, L_R$ and $L_T$ terms, which are independent from each other for better time complexity, and can be calculated with better space complexity (Section \ref{sec:space}). The $S_H$ and $S_T$ terms are just summation of the entity embedding matrix, whose calculation is even easier than the $L_H$ and $L_T$ terms. The final time complexity of non-sampling TransE is $O\left(d^2(|R|+|E|) \right)$.

\section{Experiments}
\label{sec:experiment}
In this section, we conduct experiments to evaluate both the efficiency and effectiveness of the NS-KGE framework.\footnote{Source code available at \url{https://github.com/rutgerswiselab/NS-KGE}.}

\subsection{Experimental Setup}

\subsubsection{\textbf{Dataset Description}}
We conduct the experiments on two benchmark datasets for knowledge graph embedding research, namely, FB15K237 and WN18RR. The detailed statistics of the datasets are shown in Table \ref{Table:dataset}, and we will briefly introduce these two datasets in the following.

$\textbf{FB15K237}$: One of the most frequently used dataset for KGE. The original version of the FB15K dataset is generated from a subset of the Freebase knowledge graph \cite{FB15k}. However, in the original FB15K dataset, a large number of the test triplets can be obtained by simply reversing the triplets in the training set, as shown in \cite{FB15k237,akrami2020realistic}. For example, the test set may contain a triplet (home, car, work), while the training set contains a reverse triplet (work, car, home). The existence of such cases make the original dataset suffer from the test leakage problem. As a result, the FB15k237 \cite{FB15k237} dataset is introduced by removing these reverse triplets, which mitigates the test leakage problem in a large extend. In this paper we use FB15k237 for evaluation.

$\textbf{WN18RR}$: WN18 is also a standard dataset for KGE. The original WN18 dataset is a subset of WordNet \cite{miller1995wordnet}, which is an English lexical database. Similar to FB15K, WN18 is also corrected to WN18RR \cite{WN18RR} by removing the reverse triplets to avoid test leakage. In this work, we use WN18RR for evaluation.

We use the default train-test split of the original datasets. Both the training set and the testing set are a set of $(h,r,t)$ triplets. The number of training and testing triplets of the two datasets are shown in Table \ref{Table:dataset}.

\subsubsection{\textbf{Baselines}}
We study the performance of the NS-KGE framework by comparing the performance of a KGE model with or without using the framework. Similar to what we have introduced before, we consider the following KGE models.
\begin{itemize}

\item DistMult \cite{DistMult}: The bilinear-diagonal embedding model, which uses diagonal matrix to represent the relation between head and tail entities. 

\item SimplE \cite{kazemi2018simple}: The model is a simple enhancement of the Canonical Polyadic (CP) decomposition model \cite{hitchcock1927expression} by learning two dependent embeddings for each entity.

\item ComplEx \cite{trouillon2016complex}: The model learns KG embedding in a complex number space, which uses complex number vectors to represent the entities and relations.

\item TransE \cite{bordes2013translating}: The translational embedding model, which minimizes the distance between head and tail entities after translation by the relation vector.
\end{itemize}

All of the models are implemented by PyTorch, an open source library. And we use the baselines implemented by OpenKE\footnote{https://github.com/thunlp/OpenKE} \cite{han2018openke}, an open source tool-kit for knowledge graph embedding.

\begin{table}[t]
    \centering
    \begin{tabular}{l|cccc}
        \toprule
         Dataset & \#entities & \#relations & \#train & \#test \\
         \midrule
         FB15K237 & 14,541 & 237 & 272,115 & 20,466 \\
         WN18RR & 40,943 & 11 & 86,835 & 3,134 \\
         \bottomrule
    \end{tabular}
    \caption{Basic statistics of the datasets}
    \label{Table:dataset}
    \vspace{-20pt}
\end{table}

\begin{table*}[t]
    \centering
    \setlength{\tabcolsep}{5pt}
    \begin{tabular}{l|l|l|l|l|l||l|l|l|l|l}
    \toprule
         Dataset & \multicolumn{5}{|c||}{FB15K237} & \multicolumn{5}{c}{WN18RR} \\
         \midrule
         Metric & MRR $\uparrow$ & MR $\downarrow$ & HR@10 $\uparrow$ & HR@3 $\uparrow$ & HR@1 $\uparrow$ & MRR $\uparrow$ & MR  $\downarrow$ & HR@10 $\uparrow$ & HR@3 $\uparrow$ & HR@1 $\uparrow$\\
         \midrule
         DistMult & 0.177 &430.15 &0.345 &0.198 &0.010 &0.320 & $\bm{4019.98^*}$ &0.461 &0.371 &0.240  \\
         NS-DistMult &$\bm{0.227^*}$ &$\bm{361.61^*}$ &$\bm{0.373^*}$ &$\bm{0.248^*}$ &$\bm{0.155^*}$ & $\bm{0.411^*}$ & 7456.29 & $\bm{0.462}$ & $\bm{0.424^*}$ & $\bm{0.384^*}$ \\
         \midrule
         SimplE &0.183 &387.90 & 0.355 & 0.207 & 0.099 &0.329 &$\bm{3950.82^*}$ & $\bm{0.463}$ & 0.378 & 0.253 \\
         NS-SimplE & $\bm{0.222^*}$ & $\bm{364.31^*}$ & $\bm{0.370^*}$ & $\bm{0.246^*}$ & $\bm{0.159^*}$ &$\bm{0.406^*}$ &7418.31 & 0.459 & $\bm{0.415^*}$ & $\bm{0.377^*}$ \\ 
         \midrule
         ComplEx &0.240 &529.42 &$\bm{0.415^*}$ &$\bm{0.271^*}$ &0.151 &0.390 &$\bm{4673.35^*}$ &0.474 &0.422 &0.339 \\
         NS-ComplEx & $\bm{0.243}$ &$\bm{326.57^*}$ &0.390 &0.256 &$\bm{0.163^*}$ &$\bm{0.429^*}$ &7649.34 &$\bm{0.485^*}$ &$\bm{0.449^*}$ &$\bm{0.396^*}$ \\
         \midrule
         TransE  & 0.178 & 337.86 & 0.316 & 0.192 & 0.108 & 0.079 & $\bm{2900.32^*}$ & 0.145 & 0.084 & $\bm{0.044^*}$ \\
         NS-TransE & $\bm{0.261^*}$ & $\bm{336.71}$ & $\bm{0.447^*}$ & $\bm{0.296^*}$ & $\bm{0.167^*}$ & $\bm{0.156^*}$ & 3948.07 & $\bm{0.437^*}$ & $\bm{0.256^*}$ & 0.007\\
         \bottomrule
    \end{tabular}
    \caption{Result on prediction accuracy. NS-X means the non-sampling version of model X under our NS-KGE framework. $\uparrow$ means the measure is the higher the better, while $\downarrow$ means the measure is the lower the better. Bold numbers represent better performance, and * indicates its performance is significantly better at $\bm{p < 0.01}$ than the other model.}
    \label{Table:result}
    \vspace{-15pt}
\end{table*}

\subsubsection{\textbf{Evaluation Metrics}}
For each triplet $(h,r,t)$ in the testing set, we first use $h$ and $r$ to rank all tail entities, and evaluate the position of the correct tail entity $t$. And then we use $t$ and $r$ to rank all head entities, and evaluate the position of the correct head entity $h$. As a result, if there are $|S|$ triplets in the testing set, we will conduct $2|S|$ evaluations.

We use Hit Ratio ($\bm {HR}$), Mean Rank ($\bm {MR}$) and Mean Reciprocal Rank ($\bm {MRR}$) to evaluate the models. HR is used  to measure whether the correct entity is in the Top-K list. MR is the mean of the correct entity's rank, defined as $MR=\frac{1}{2|S|}\sum_{(h,r,t)\in S}(rank_h+rank_t)$.
MRR is defined as  $MRR=\frac{1}{2|S|}\sum_{(h,r,t)\in S}(\frac{1}{rank_h}+\frac{1}{rank_t})$. These three metrics are widely used in KG embedding evaluation \cite{bordes2013translating, lin2015learning, kazemi2018simple}. For HR and MRR, larger value means better performance, and for MR, smaller value means better performance.

\subsubsection{\textbf{Parameter Settings}}
We set the default embedding dimension as 200, the number of training epochs as 2000, initial learning rate as 0.0001, and use Adam optimizer \cite{kingma2015adam} for all models. To avoid over-fitting, we apply $\ell_2$ normalization over the parameters for all models, and we conduct grid search to find the best coefficient of regularization for each model in  $[10^{-1},\ 10^{-2},\ 10^{-3},\ 10^{-4}]$. We also conduct grid search to find the best learning rate decay for each model in $[0.1,\ 0.3,\ 0.5,\ 0.7]$.

For negative sampling-based models, 
we set the number of negative samples as 25; the batch size is 4000. For non-sampling models, we do not split training data into batches, because our model has lower space complexity; the coefficient of positive instances $c^+$ is set to 1, and the coefficient of negative instances $c^-$ is grid searched in $[10^{-1},10^{-2},10^{-3},\ 10^{-4},\  10^{-5},\ 10^{-6}]$. The default setting of $c^-$ is 0.001 in all experiments except when we are tuning $c^-$ to see its influence.
For each model on each dataset, we run the model 5 times and report the average result of the 5 times. We use paired $t$-test to verify the significance of the results.

\begin{table*}[t]
    \centering
    \setlength{\tabcolsep}{5pt}
    \begin{tabular}{l|l|l|l|l}
    \toprule
         Dataset & \multicolumn{4}{|c}{FB15K237} \\
         \midrule
         Relation & \multicolumn{4}{|c}{media common/netflix genre/titles} \\
         \midrule
         Entity & \multicolumn{2}{|c|}{The Notebook} &\multicolumn{2}{|c}{Funny Girl}  \\
         \midrule
         Model & NS-DistMult  & DistMult & NS-SimplE & SimplE  \\
         \midrule
         \multirow{10}{0.08\textwidth}{\centering Predicted Top-10 Entities}
          & \textcolor{gray}{comedy} & \textcolor{gray}{comedy} &comedy &United States of America   \\
          &\textbf{historical period drama} & \textcolor{gray}{drama} &\textbf{historical period drama} & \textcolor{gray}{drama}  \\
         
          & \textcolor{gray}{fantasy} & \textcolor{gray}{thriller} & fantasy & romance film \\
          & \textcolor{gray}{drama}  &romance film & \textcolor{gray}{drama} & \textcolor{gray}{thriller}  \\ 
         
          &biography  & \textcolor{gray}{musical film} & musical film & \textcolor{gray}{psychological thriller}   \\
          & \textcolor{gray}{thriller} & \textcolor{gray}{fantasy} &biography &DVD  \\
          & \textcolor{gray}{musical film} & \textcolor{gray}{suspense} &political drama &\textbf{historical period drama}  \\
          & psychological thriller &\textbf{historical period  drama} & \textcolor{gray}{psychological thriller} &crime fiction  \\
          & \textcolor{gray}{suspense} & \textcolor{gray}{mystery} & \textcolor{gray}{thriller} &mystery fiction  \\
          & \textcolor{gray}{mystery} &United States of America &suspense &United Kingdom  \\
         
         
         \bottomrule
    \end{tabular}
    \caption{Qualitative results on entity ranking. NS-X means the non-sampling version of model X under our NS-KGE framework. Bold entities are the ground truth. Common entities between model X and NS-X are in gray to highlight the difference entities.}
    \label{Table:qualitative_result}
    \vspace{-20pt}
\end{table*}

\subsection{Performance Comparison}

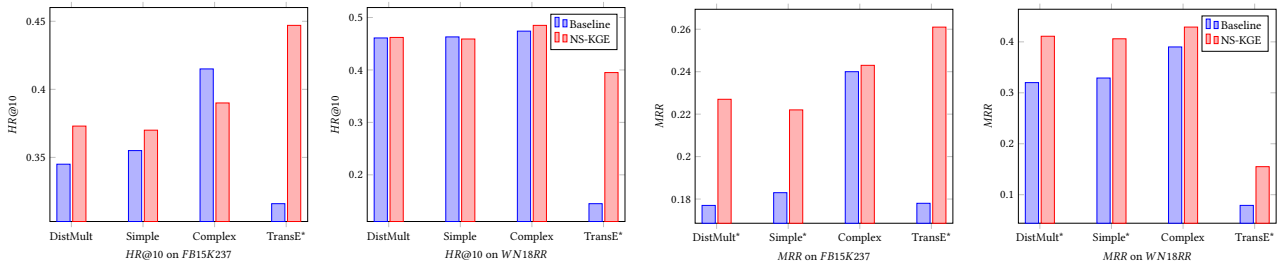
\begin{figure*}[t]

\subfigure{
\begin{minipage}[t]{0.25\linewidth}
\centering
\pgfplotstableread[row sep=\\,col sep=&]{
    interval & neg & non  \\
    DistMult     & 0.345  & 0.373    \\
    Simple     & 0.355 & 0.370    \\
    Complex    & 0.415 & 0.390  \\
    TransE*     & 0.316 & 0.447  \\
   }\mydata
\begin{tikzpicture}[scale=0.5]
    \begin{axis}[
            ylabel near ticks,
            ybar,
            symbolic x coords={DistMult,Simple,Complex,TransE*},
            xtick=data,
             xlabel=$HR@10$ on $FB15K237$,
             ylabel=$HR@10$,
            legend pos=north east
        ]
        \addplot table[x=interval,y=neg]{\mydata};
        \addplot table[x=interval,y=non]{\mydata};

    \end{axis}
\end{tikzpicture}%
\end{minipage}%
}%
\hspace{-10pt}
\subfigure{
\begin{minipage}[t]{0.25\linewidth}
\centering
\pgfplotstableread[row sep=\\,col sep=&]{
    interval & neg & non  \\
    DistMult     & 0.461  & 0.462    \\
    Simple     & 0.463 & 0.459    \\
    Complex    & 0.474 & 0.485  \\
    TransE*    & 0.145  & 0.395 \\
    }\mydata%
\begin{tikzpicture}[scale=0.5]
    \begin{axis}[
            ylabel near ticks,
            ybar,
            symbolic x coords={DistMult,Simple,Complex,TransE*},
            xtick=data,
             xlabel=$HR@10$ on $WN18RR$,
             ylabel=$HR@10$,
            legend pos=north east
        ]
        \addplot table[x=interval,y=neg]{\mydata};
        \addplot table[x=interval,y=non]{\mydata};
         \legend{Baseline, NS-KGE}

    \end{axis}
\end{tikzpicture}
\end{minipage}%
}
\hspace{-10pt}
%
\subfigure{
\begin{minipage}[t]{0.25\linewidth}
\centering
\pgfplotstableread[row sep=\\,col sep=&]{
    interval & neg & non  \\
    DistMult*     & 0.177  & 0.227    \\
    Simple*     & 0.183 & 0.222    \\
    Complex    & 0.240 & 0.243  \\
   TransE*     & 0.178 & 0.261  \\
    }\mydata
    
\begin{tikzpicture}[scale=0.5]
    \begin{axis}[
            ylabel near ticks,
            ybar,
            symbolic x coords={DistMult*,Simple*,Complex,TransE*},
            xtick=data,
             xlabel=$MRR$ on $FB15K237$,
             ylabel=$MRR$,
            legend pos=north west
        ]
        \addplot table[x=interval,y=neg]{\mydata};
        \addplot table[x=interval,y=non]{\mydata};

    \end{axis}
\end{tikzpicture}%
\end{minipage}%
}
\hspace{-10pt}
\subfigure{
\begin{minipage}[t]{0.25\linewidth}
\centering
\pgfplotstableread[row sep=\\,col sep=&]{
    interval & neg & non  \\
    DistMult*     & 0.320  & 0.411    \\
    Simple*     & 0.329 & 0.406    \\
    Complex    & 0.390 & 0.429  \\
    TransE*     & 0.079 & 0.155  \\
    }\mydata%
\begin{tikzpicture}[scale=0.5]
    \begin{axis}[
            ylabel near ticks,
            ybar,
            symbolic x coords={DistMult*,Simple*,Complex,TransE*},
            xtick=data,
             xlabel=$MRR$ on $WN18RR$,
             ylabel=$MRR$,
            legend pos=north east
        ]
        \addplot table[x=interval,y=neg]{\mydata};
        \addplot table[x=interval,y=non]{\mydata};
         \legend{Baseline, NS-KGE}
    \end{axis}
\end{tikzpicture}
\end{minipage}%
}%
\vspace{-15pt}
 \caption{Performance on HR@10 and MRR with and without the NS-KGE framework. The models on x-axis labeled with ``*'' mean that the performance of NS-KGE framework improved more than 20\% from the corresponding baselines.}
 \label{fig:performance}
\vspace{-5pt}
\end{figure*}

We apply NS-KGE to DistMult, SimplE, ComplEx and TransE. 
The experimental results on prediction accuracy are shown in Table \ref{Table:result}, and more intuitive comparison are shown in Figure \ref{fig:performance}. We have the following observations from the results.

First and most importantly, compared to the four baselines, in most cases, our NS-KGE framework achieves the best performance on both of the two datasets. Although some baselines are slightly better than NS-KGE in some cases, for example, on the WN18RR dataset, SimplE's HR@10 has a slightly better performance, but we can see the results are comparable. For HR@1 and HR@3, NS-KGE has 9.78\% and 49.01\% improvement on average, respectively.

We also conducted some qualitative analysis of the entity ranking results, as shown in Table \ref{Table:qualitative_result}. First, for the same entity and relation, we see that the correct prediction gains higher rank in our non-sampling models. Second, compared to the sampling-based model, the top-10 ranked entities by our non-sampling model tend to be intuitively more relevant to the given entity and relation.


\begin{figure*}
\centering
\subfigure{
\begin{minipage}[t]{0.25\linewidth}
\centering
\begin{tikzpicture}[scale=0.5] 
    \begin{axis}[
        ymin=0.3,
        ylabel near ticks,
        xlabel near ticks,
        xlabel=$HR@10$ on $FB15K237$,
        ylabel=$HR@10$,
        legend pos=south east]
    \addplot[smooth,mark=*,blue,dash pattern=on 1pt off 3pt on 3pt off 3pt] plot coordinates {
        (50,0.352)
        (100,0.364)
        (150,0.369)
        (200,0.374)
        (250,0.374)
        (300,0.377)
        (350,0.382)
        (400,0.383)
    };
    
    \addlegendentry{NS-Simple}
    \addplot[smooth,mark=square*,red,dash pattern=on 3pt off 6pt on 6pt off 6pt] plot coordinates {
        (50,0.363)
        (100,0.374)
        (150,0.380)
        (200,0.384) 
        (250,0.390)
        (300,0.390)
        (350,0.395)
        (400,0.397)
    };
    \addlegendentry{NS-Complex}   
    
    \addplot[smooth,mark=diamond*,olive,densely dotted] plot coordinates {
        (50,0.350)
        (100,0.363)
        (150,0.371)
        (200,0.375)
        (250,0.384)
        (300,0.385)
        (350,0.388)
        (400,0.394)
    };
    \addlegendentry{NS-DistMult}   
    
    \addplot[smooth,mark=triangle*,black] plot coordinates {
        (50,0.396)
        (100,0.427)
        (150,0.441)
        (200,0.447)
        (250,0.449)
        (300,0.450)
        (350,0.450)
        (400,0.449)
    };
    \addlegendentry{NS-TransE}

    \end{axis}
    \end{tikzpicture}
    \end{minipage}%
}%
\hspace{-10pt}
\subfigure{
\begin{minipage}[t]{0.25\linewidth}
\centering
\begin{tikzpicture}[scale=0.5] 
    \begin{axis}[
        ylabel near ticks,
        xlabel near ticks,
        xlabel=$HR@10$ on $WN18RR$,
        ylabel=$HR@10$,
        legend pos=south east]
    \addplot[smooth,mark=*,blue,dash pattern=on 1pt off 3pt on 3pt off 3pt] plot coordinates { 
        (50,0.367) 
        (100,0.426)
        (150,0.447)
        (200,0.451) 
        (250,0.463)
        (300,0.471)
        (350,0.477)
        (400,0.478)
    };
    
    \addplot[smooth,mark=square*,red,dash pattern=on 3pt off 6pt on 6pt off 6pt] plot coordinates {
        (50,0.421)
        (100,0.461)
        (150,0.480)
        (200,0.484)
        (250,0.486)
        (300,0.495)
        (350,0.500)
        (400,0.503)
    };
    
    \addplot[smooth,mark=diamond*,olive,densely dotted] plot coordinates {
        (50,0.381)
        (100,0.440)
        (150,0.462)
        (200,0.474)
        (250,0.486)
        (300,0.487)
        (350,0.491)
        (400,0.494)
    };
    
    \addplot[smooth,mark=triangle*,black] plot coordinates {
        (50,0.4)
        (100,0.433)
        (150,0.434)
        (200,0.434)
        (250,0.433)
        (300,0.433)
        (350,0.431)
        (400,0.431)
    };

    \end{axis}
     
    \end{tikzpicture}%
\end{minipage}%
}%
\hspace{-10pt}
\subfigure{
\begin{minipage}[t]{0.25\linewidth}
\centering
\begin{tikzpicture}[scale=0.5] 
    \begin{axis}[
        ymin=0.2,
        ylabel near ticks,
        xlabel near ticks,
        xlabel=$MRR$ on $FB15K237$,
        ylabel=$MRR$,
        legend pos=south east]
    \addplot[smooth,mark=*,blue,dash pattern=on 1pt off 3pt on 3pt off 3pt] plot coordinates {
        (50,0.223)
        (100,0.225)
        (150,0.226)
        (200,0.231)
        (250,0.232)
        (300,0.232)
        (350,0.239)
        (400,0.239)
    };
    
    \addlegendentry{NS-Simple}
    \addplot[smooth,mark=square*,red,dash pattern=on 3pt off 6pt on 6pt off 6pt] plot coordinates {
        (50,0.221)
        (100,0.229)
        (150,0.233)
        (200,0.237) 
        (250,0.240)
        (300,0.242)
        (350,0.243)
        (400,0.243)
    };
    \addlegendentry{NS-Complex}   
    
    \addplot[smooth,mark=diamond*,olive,densely dotted] plot coordinates {
        (50,0.227)
        (100,0.231)
        (150,0.233)
        (200,0.235)
        (250,0.242)
        (300,0.242)
        (350,0.244)
        (400,0.247)
    };
    \addlegendentry{NS-DistMult}     
    
    \addplot[smooth,mark=triangle*,black] plot coordinates {
        (50,0.236)
        (100,0.250)
        (150,0.257)
        (200,0.262)
        (250,0.264)
        (300,0.265)
        (350,0.265)
        (400,0.266)
    };
    \addlegendentry{NS-TransE}    

    \end{axis}
    \end{tikzpicture}
\end{minipage}%
}%
\hspace{-5pt}
\subfigure{
\begin{minipage}[t]{0.25\linewidth}
\centering
\begin{tikzpicture}[scale=0.5] 
    \begin{axis}[
        ylabel near ticks,
        xlabel near ticks,
        xlabel=$MRR$ on $WN18RR$,
        ylabel=$MRR$,
        legend pos=south east]
    \addplot[smooth,mark=*,blue,dash pattern=on 1pt off 3pt on 3pt off 3pt] plot coordinates { 
        (50,0.298) 
        (100,0.393)
        (150,0.409)
        (200,0.411) 
        (250,0.420)
        (300,0.423)
        (350,0.427)
        (400,0.426)
    };
    
    \addplot[smooth,mark=square*,red,dash pattern=on 3pt off 6pt on 6pt off 6pt] plot coordinates {
        (50,0.349)
        (100,0.407)
        (150,0.426)
        (200,0.428)
        (250,0.433)
        (300,0.426)
        (350,0.438)
        (400,0.439)
    };
    
    \addplot[smooth,mark=diamond*,olive,densely dotted] plot coordinates {
        (50,0.300)
        (100,0.394)
        (150,0.413)
        (200,0.422)
        (250,0.428)
        (300,0.431)
        (350,0.433)
        (400,0.434)
    };
    
    \addplot[smooth,mark=triangle*,black] plot coordinates {
        (50,0.144)
        (100,0.155)
        (150,0.156)
        (200,0.155)
        (250,0.154)
        (300,0.153)
        (350,0.151)
        (400,0.150)
    };

    \end{axis}
     
    \end{tikzpicture}%
\end{minipage}%
}
\vspace{-15pt}
\caption{Performance on HR@10 (left two figures) and MRR (right two figures) under different dimension size $d$.}
\vspace{-10pt}
\label{figure:dimension}
\end{figure*}
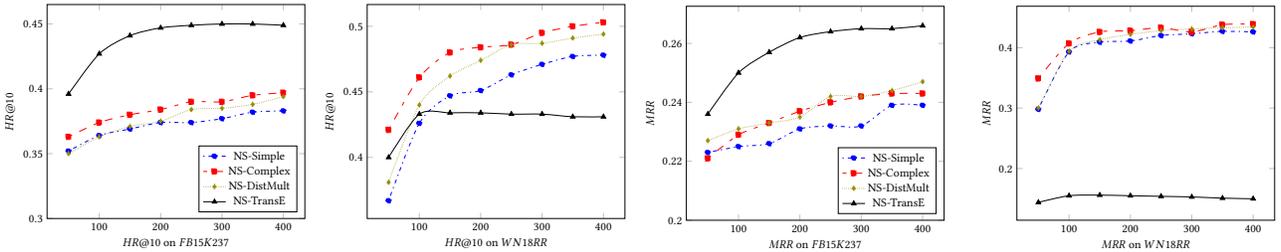

The reason why applying the NS-KGE framework can improve the performance of sampling-based methods (DistMult, SimplE, Complex and TransE) is that, the sampling-based methods only use part of the negative instance information in the dataset, and these models may ignore some important negative instances.
However, our NS-KGE framework makes it possible to use all of the information in the dataset and brings better computational time and space consumption at the same time(to be discussed in the following experiments). Therefore, NS-KGE can avoid the problem of sampling-based methods and thus improve the performance.

One interesting observation is that on the WN18RR dataset, our NS-KGE framework is consistently better on the MRR measure, but is worse on the MR measure. The difference between MRR and MR is that MR is more sensitive to bad cases. Due to the large number of entities in the WN18RR dataset, if the correct entity is ranked to lower positions in some lists, it will have a huge influence on the MR measure, but not too much on the MRR measure due to the reciprocal operation. The result implies that our framework may rank the correct entity to very low positions in some cases. However, since our performance on MRR is better, it means that in most cases our framework ranks the correct entity to top positions.

Besides, the observation that NS-KGE improves the performance of both factorization-based (DistMult, SimplE, ComplEx) models and translation-based (TransE) models indicates 
the effectiveness of the NS-KGE framework, and also shows the potential of applying the framework on other KG embedding models.

\begin{figure*}[t]
\centering
\subfigure{
\begin{minipage}[t]{0.25\linewidth}
\centering
\begin{tikzpicture}[scale=0.5] 
    \begin{axis}[
        ymin=0,
        ylabel near ticks,
        xlabel near ticks,
        xlabel=$HR@10$ on $FB15K237$,
        ylabel=$HR@10$,
        xtick = {1,2,3,4,5,6},
  xticklabels = {0.000001,0.00001,0.0001,0.001,0.01,0.1},
        legend pos= south west]
    \addplot[smooth,mark=*,blue,dash pattern=on 1pt off 3pt on 3pt off 3pt] plot coordinates {
        (1,0.326)
        (2,0.344)
        (3,0.374) 
        (4,0.372)
        (5,0.355)
        (6,0.295)
    };
    
    \addlegendentry{NS-Simple}
    \addplot[smooth,mark=square*,red,dash pattern=on 3pt off 6pt on 6pt off 6pt] plot coordinates {
        (1,0.345)
        (2,0.363)
        (3,0.384)
        (4,0.38)
        (5,0.37)
        (6,0.345)
    };
    \addlegendentry{NS-Complex}   
    
    \addplot[smooth,mark=diamond*,olive,densely dotted] plot coordinates {
        (1,0.339)
        (2,0.355)
        (3,0.372) 
        (4,0.371)
        (5,0.360)
        (6,0.345)
    };
    \addlegendentry{NS-DistMult}    
    
    \addplot[smooth,mark=triangle*,black] plot coordinates {
        (1,0.363)
        (2,0.418)
        (3,0.447)
        (4,0.361)
        (5,0.208)
        (6,0.090)
    };
    \addlegendentry{NS-TransE}    

    \end{axis}
    \end{tikzpicture}
\end{minipage}%
}%
\hspace{-10pt}%
\subfigure{
\begin{minipage}[t]{0.25\linewidth}
\centering
\begin{tikzpicture}[scale=0.5] 
    \begin{axis}[
        ymin=0,
        ylabel near ticks,
        xlabel near ticks,
        xlabel=$Hit@10$ on $WN18RR$,
        ylabel=$Hit@10$,
        xtick = {1,2,3,4,5,6},
  xticklabels = {0.000001,0.00001,0.0001,0.001,0.01,0.1},
        ]
    \addplot[smooth,mark=*,blue,dash pattern=on 1pt off 3pt on 3pt off 3pt] plot coordinates { 
        (1,0.442)
        (2,0.441)
        (3,0.456)
        (4,0.457)
        (5,0.351)
        (6,0.201)
    };
    
    \addplot[smooth,mark=square*,red,dash pattern=on 1pt off 6pt on 6pt off 6pt] plot coordinates {
        (1,0.166)
        (2,0.158)
        (3,0.480)
        (4,0.50)
        (5,0.46)
        (6,0.30)
        
    };
    
    \addplot[smooth,mark=diamond*,olive,densely dotted] plot coordinates {
        (1,0.442)
        (2,0.450)
        (3,0.466)
        (4,0.47)
        (5,0.382)
        (6,0.21)
    };

    \addplot[smooth,mark=triangle*,black] plot coordinates {
        (1,0.156)
        (2,0.303)
        (3,0.414)
        (4,0.437)
        (5,0.409)
        (6,0.013)
    };
    \end{axis}
     
    \end{tikzpicture}%
    \end{minipage}%
}%
\hspace{-10pt}%
\subfigure{
\begin{minipage}[t]{0.25\linewidth}
\centering
\begin{tikzpicture}[scale=0.5] 
    \begin{axis}[
        ymin=0,
        ylabel near ticks,
        xlabel near ticks,
        xlabel=$MRR$ on $FB15K237$,
        ylabel=$MRR$,
        ymin = 0,
        ymax = 0.4,
        xtick = {1,2,3,4,5,6},
  xticklabels = {0.000001,0.00001,0.0001,0.001,0.01,0.1},
        legend pos= south west
        ]
    \addplot[smooth,mark=*,blue,dash pattern=on 1pt off 3pt on 3pt off 3pt] plot coordinates {
        (1,0.187)
        (2,0.202)
        (3,0.228) 
        (4,0.238)
        (5,0.230)
        (6,0.194)
    };
    
    \addlegendentry{NS-Simple}
    \addplot[smooth,mark=square*,red,dash pattern=on 3pt off 6pt on 6pt off 6pt] plot coordinates {
        (1,0.200)
        (2,0.213)
        (3,0.237)
        (4,0.24)
        (5,0.242)
        (6,0.22)
    };
    \addlegendentry{NS-Complex}   
    
    \addplot[smooth,mark=diamond*,olive,densely dotted] plot coordinates {
        (1,0.196)
        (2,0.210)
        (3,0.230)
        (4,0.236)
        (5,0.23)
        (6,0.20)
    };
    \addlegendentry{NS-DistMult}    
    
    \addplot[smooth,mark=triangle*,black] plot coordinates {
        (1,0.219) 
        (2,0.248)
        (3,0.261)
        (4,0.196) 
        (5,0.085)
        (6,0.037)
    };
    \addlegendentry{NS-TransE}    

    \end{axis}
    \end{tikzpicture}
    \end{minipage}
}%
\hspace{-10pt}%
\subfigure{
\begin{minipage}[t]{0.25\linewidth}
\centering
\begin{tikzpicture}[scale=0.5] 
    \begin{axis}[
        ymin=0,
        ylabel near ticks,
        xlabel near ticks,
        xlabel=$MRR$ on $WN18RR$,
        ylabel=$MRR$,
        xtick = {1,2,3,4,5,6},
  xticklabels = {0.000001,0.00001,0.0001,0.001,0.01,0.1},
        ]
    \addplot[smooth,mark=*,blue,dash pattern=on 1pt off 3pt on 3pt off 3pt] plot coordinates { 
        (1,0.408)
        (2,0.408)
        (3,0.417) 
        (4,0.40)
        (5,0.27)
        (6,0.12)
    };
    
    \addplot[smooth,mark=square*,red,dash pattern=on 1pt off 6pt on 6pt off 6pt] plot coordinates {
        (1,0.166)
        (2,0.158)
        (3,0.430)
        (4,0.45)
        (5,0.37)
        (6,0.20)
    };
    
    \addplot[smooth,mark=diamond*,olive,densely dotted] plot coordinates {
        (1,0.409)
        (2,0.410)
        (3,0.423) 
        (4,0.41)
        (5,0.289)
        (6,0.135)
    };

    \addplot[smooth,mark=triangle*,black] plot coordinates {
        (1,0.084) 
        (2,0.120)
        (3,0.149)
        (4,0.156)
        (5,0.149)
        (6,0.036)
    };
    \end{axis}
     
    \end{tikzpicture}%
\end{minipage}%
}%
\vspace{-15pt}
\caption{Performance on HR@10 (left two figures) and MRR (right two figures) under different negative instance weight $c^-$.}
\label{figure: negative}
\vspace{0pt}
\end{figure*}
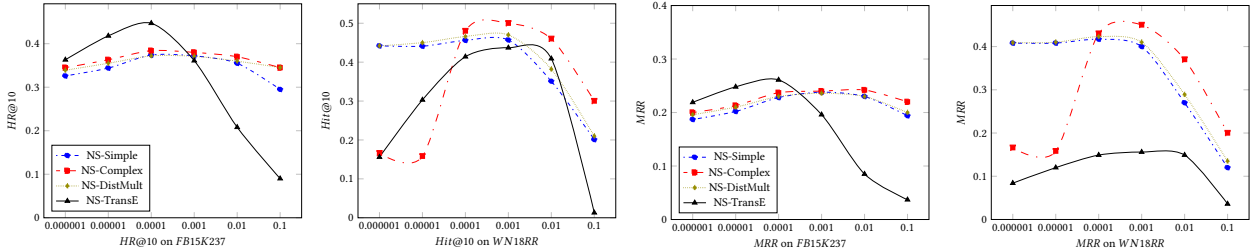

\subsection{Analyses of Hyper Parameters}
In this section, we will analyze the impact of different dimension size $d$ and the negative instance weight $c^-$.

%
%
%
%

\textbf{Impact of dimension size}. Figure \ref{figure:dimension} shows the performance of our NS-KGE framework under different choices of embedding dimension size $d$. We can see that in most cases, the performance becomes better when the embedding dimension size increases. It indicates that higher model expressiveness power contributes to better performance in the NS-KGE framework. 
However, a larger dimension size can also cost more computing time in model training. As we have shown in Section \ref{section: efficiency}, the training time is proportional to the square of dimension size. Therefore, we need to trade-off between the training time and the performance. As we can see in Figure \ref{figure:dimension}, in most cases, the performance tends to be stable at round 200 dimensions. As a result, we choose 200 as the default dimension size for all of the models.



\textbf{Impact of negative instance weight}. In this experiment, we fix the positive instance weight in the NS-KGE framework as $c^+=1$, and we tune the negative instance weight $c^-$ to analyze its influence. Figure \ref{figure: negative} shows the results when we change the negative instance weight $c^-$ on the two datasets. We see that for all models on both datasets, when the value of $c^-$ increases, the performance tends to increase first and then decreases when $c^-$ is too large. 
This shows that a proper selection of $c^-$ value is important to the model performance. If $c^-$ is too small (e.g., close to 0), the model would not be able to leverage the information included in the negative instances of the KG. However, the information in negative instances is also noisy, e.g., if two entities are not connected, this may not directly indicate they are irrelevant, instead, this may be caused by the noise in data collection process. As a result, negative instance information is not as reliable as positive instances, and if $c^-$ is too large, it may decrease the performance. Because negative samples are usually much more than positive samples, to avoid class imbalance, the weight of negative instances $c^-$ should be smaller than $c^+$. In most cases, the optimal selection of $c^-$ is 0.001. 



\subsection{Efficiency Analyses}
\label{experiment:efficiency}


In this section, we will discuss the training efficiency of our NS-KGE framework. We will compare the training time of the four sampling-based models DistMult, SimplE, Complex, TransE and their non-sampling versions. For fairness of comparison, all experiments run on a single NVIDIA Geforce 2080Ti GPU. The operating system is Ubuntu 16.04 LTS. For all models, the embedding dimension is set as 200 and number of training epochs is 2000. Results on model training time is shown in Table \ref{Table:running time}.

\begin{table}[hbt]
    \centering
    \setlength{\tabcolsep}{5pt}
    \begin{tabular}{l|rr|rr}
        \toprule
         & FB15K237 & Speed-up & WN18RR & Speed-up \\
         \midrule
         DistMult &3546s & 1.00 &1922s & 1.00 \\
         NS-DistMult & \textbf{53s} & 66.91 & \textbf{57s} & 33.72 \\
         \midrule
         SimplE & 4447s & 1.00 &2450s & 1.00\\
         NS-SimplE & \textbf{73s} & 60.92 & \textbf{77s} & 31.82 \\
         \midrule
         TransE & 2353s & 1.00 & 673s & 1.00 \\
         NS-TransE & \textbf{107s} & 21.99 & \textbf{86s} & 7.83\\
         \midrule
         ComplEx &6736s & 1.00 &3346s & 1.00 \\
         NS-ComplEx & \textbf{157s} & 42.90 & \textbf{158s} & 21.18 \\
         \bottomrule
    \end{tabular}
    \caption{Experimental results on model training time. The models are ordered from top to bottom in ascending order of the training time on each dataset. Speed-up shows how many times NS-X is faster than the corresponding model X.}
    \vspace{-15pt}
    \label{Table:running time}
\end{table}

We can see that the training efficiency of our NS-KGE framework is significant better than the baseline models. 
For example, if we apply NS-KGE to the DistMult model on the FB15K237 dataset, it only takes 53s to finish training the model, while the original sampling-based DistMult model takes 3546s. The acceleration is about 70 times. For other models and datasets, we also get $20\sim60$ times acceleration. This is not surprising because for the sampling-based KGE models, a lot of computational time needs to be spent on sampling the negative examples, while our framework eliminates the sampling procedure. In our implementation, we used in-memory sampling instead of on-disk sampling for the baselines, however, our NS-KGE framework is still much faster than the baselines.


Another intersting observation from Table \ref{Table:running time} is that on each dataset, the computational time of our NS-KGE models are NS-DistMult < NS-SimplE < NS-TransE < NS-ComplEx. This is consistent with the mathematical analysis in Section \ref{sec:framework} and Section \ref{sec:application}. For the NS-DistMult model, its final loss function has one $L^A$ term (see Eq.\eqref{equation:3.3final}). For the NS-SimplE model, its final loss function has three $L^A$ terms (Eq.\eqref{equation:11}). For the NS-TransE model, its final loss function has four $L^A$ terms (Eq.\eqref{equation:transe}). While for the NS-ComplEx model, it final loss function has six $L^A$ terms (see Section \ref{section:complex}, we only show one of terms in Eq.\eqref{equation:13}). As we have shown in Section \ref{section: efficiency}, the $L^A$ term(s) take the most significant computational time in the loss function. As a result, the final model training time is proportional to the number of $L^A$ terms in the loss function.



\section{Conclusions and Future Work}
\label{sec:conclusions}
In this paper, we proposes NS-KGE, a non-sampling framework for knowledge graph embedding, which leverages all of the positive and negative instances in the KG for model training. Besides, we provided mathematical methods to reduce the time and space complexity of the framework, and have shown how the framework can be applied to various KGE models. Experiments on two benchmark datasets demonstrate that the framework is able to enhance both the model performance and the training efficiency.

In the future, we will consider applying our framework on more complex KGE models such as neural network-based models, as well as more complex graph structures such as attributed graphs. We also plan to apply our framework to other graph computation models beyond KGE, such as graph neural networks.

\section*{Acknowledgement}
We appreciate the reviews and suggestions of the reviewers. This work was supported in part by NSF IIS-1910154 and IIS-2007907. Any opinions, findings, conclusions or recommendations expressed in this material are those of the authors and do not necessarily reflect those of the sponsors.

\bibliographystyle{unsrt}
\bibliography{paper.bib}

\end{document}